\documentclass{article}
 
\usepackage{microtype}
\usepackage{graphicx}
\usepackage{booktabs} % for professional tables
\usepackage{comment}
\usepackage{subcaption}
\usepackage{multirow}

\usepackage{hyperref}

\usepackage[accepted]{icml2019}

\usepackage{array}
\newcolumntype{?}{!{\vrule width 1pt}}

\usepackage{xcolor}

\usepackage{amsmath,amsfonts}
\usepackage{amssymb,amsopn}
\usepackage{bm} % bold symbol
\usepackage{dsfont}
\newcommand{\vct}[1]{\boldsymbol{\mathbf{#1}}} % vector
\newcommand{\mat}[1]{\boldsymbol{\mathbf{#1}}} % matrix
\newcommand{\x}{\vct{x}} % vector x
\newcommand{\z}{\vct{z}} % vector z
\newcommand{\h}{\vct{h}} % vector z
\newcommand{\W}{\mat{W}} % matrix W
\newcommand{\I}{\mat{I}} % matrix W
\newcommand{\vtheta}{\vct{\theta}} % vector \theta
\newcommand{\vphi}{\vct{\phi}} % vector \phi
\newcommand{\relu}{\text{ReLU}} % ReLU activation

\DeclareRobustCommand\onedot{\futurelet\@let@token\@onedot}
\def\onedot{. } %\@\xspace}

\def\eg{\textit{e.g}\onedot} 
\def\ie{\textit{i.e}\onedot}

\icmltitlerunning{Variational Laplace Autoencoders}

\begin{document}

\twocolumn[
\icmltitle{Variational Laplace Autoencoders}

% It is OKAY to include author information, even for blind
% submissions: the style file will automatically remove it for you
% unless you've provided the [accepted] option to the icml2019
% package.

% List of affiliations: The first argument should be a (short)
% identifier you will use later to specify author affiliations
% Academic affiliations should list Department, University, City, Region, Country
% Industry affiliations should list Company, City, Region, Country

% You can specify symbols, otherwise they are numbered in order.
% Ideally, you should not use this facility. Affiliations will be numbered
% in order of appearance and this is the preferred way.
\icmlsetsymbol{equal}{*}

\begin{icmlauthorlist}
\icmlauthor{Yookoon Park}{to}
\icmlauthor{Chris Dongjoo Kim}{to}
\icmlauthor{Gunhee Kim}{to}
\end{icmlauthorlist}

\icmlaffiliation{to}{Neural Processing Research Center, Seoul National University, Seoul, South Korea}

\icmlcorrespondingauthor{Gunhee Kim}{gunhee@snu.ac.kr}

% You may provide any keywords that you
% find helpful for describing your paper; these are used to populate
% the "keywords" metadata in the PDF but will not be shown in the document
\icmlkeywords{Variational Autoencoders, Variational Inferece, Deep Generative Models, Amortization Gap, Approximation Gap}

\vskip 0.3in
]

% this must go after the closing bracket ] following \twocolumn[ ...

% This command actually creates the footnote in the first column
% listing the affiliations and the copyright notice.
% The command takes one argument, which is text to display at the start of the footnote.
% The \icmlEqualContribution command is standard text for equal contribution.
% Remove it (just {}) if you do not need this facility.

\printAffiliationsAndNotice{}  % leave blank if no need to mention equal contribution
% \printAffiliationsAndNotice{\icmlEqualContribution} % otherwise use the standard text.

\begin{abstract}
  Variational autoencoders \cite{kingma14} employ an amortized inference model to approximate the posterior of latent variables. However, such amortized variational inference faces two challenges: (1) the limited posterior expressiveness of fully-factorized Gaussian assumption and (2) the amortization error of the inference model. 
  We present a novel approach that addresses both challenges. 
  First, we focus on ReLU networks with Gaussian output and illustrate their connection to probabilistic PCA. Building on this observation, we derive an iterative algorithm that finds the mode of the posterior and apply full-covariance Gaussian posterior approximation centered on the mode. 
  %  We propose an extended model named \textit{Variational Laplace Autoencoders} that overcome both challenges to improve the training of deep generative models. Specifically, we start from a class of neural networks with rectified linear activations and Gaussian output, and create a connection to probabilistic PCA.  As a result, we derive iterative update equations that discover the mode of the posterior and define a local full-covariance Gaussian approximation centered on it. 
  Subsequently, we present a general framework named \textit{Variational Laplace Autoencoders} (VLAEs) for training deep generative models. Based on the \textit{Laplace approximation} of the latent variable posterior, VLAEs enhance the expressiveness of the posterior while reducing the amortization error. 
  %  From the perspective of Laplace approximation, we generalize the model to a differentiable class of output distributions and activation functions. 
  Empirical results on MNIST, Omniglot, Fashion-MNIST, SVHN and CIFAR10 show that the proposed approach significantly outperforms other recent amortized or iterative methods on the ReLU networks. %\protect\footnotemark
\end{abstract}

\section{Introduction}
\label{sec:intro}
Variational autoencoders (VAEs) \cite{kingma14} are deep latent generative models which have been popularly used in various domains of data such as images, natural language and sound \cite{gulrajani17, gregor15, bowman15, chung15, roberts18}. VAEs introduce an amortized inference network (\ie an \textit{encoder}) to approximate the posterior distribution of latent variable $\z$ and maximize the evidence lower-bound (ELBO) of the data.
However, the two major limitations of VAEs are: (1) the constrained expressiveness of the fully-factorized Gaussian posterior assumption and (2) the amortization error \cite{cremer18} of the inference model due to the dynamic posterior prediction. 

There have been various attempts to address these problems. A representative line of works belongs to the category of normalizing flows \cite{rezende15, kingma16, tomczak16}, which apply a chain of invertible transformation with tractable densities in order to represent a more flexible posterior distribution. 
However, not only do they incur additional parameter overhead for the inference model but are yet prone to the amortization error as they entirely depend on the dynamic inference. 

Recently, iterative approaches based on  gradient-based refinement of the posterior parameters have been proposed \cite{krishnan18, kim18, marino18}.
These methods aim to reduce the amortization error by augmenting the dynamic inference with an additional inner-loop optimization of the posterior. 
Nonetheless, they still rely on the fully-factorized Gaussian assumption, and accordingly fail to enhance the expressiveness of the posterior.

We develop a novel approach that tackles both challenges by (1) iteratively updating the mode of the approximate posterior and (2) defining a full-covariance Gaussian posterior centered on the mode, whose covariance is determined by the local behavior of the generative network. By deducing the approximate posterior directly from the generative network, not only are we able to minimize the amortization error but also model the rich correlations between the latent variable dimensions. 

We start from the class of neural networks of rectified linear activations (\textit{e.g.} ReLU) \cite{montufar14, pascanu14} and Gaussian output, which is universally popular for modeling continuous data such images \cite{krizhevsky12, gregor15} and also as a building block of deep latent models \cite{rezende14, sonderby16}. 
Subsequently, we present a generalized framework named \textit{Variational Laplace Autoencoders} (VLAEs), which encompasses the general class of differentiable neural networks. 
We show that the ReLU networks are in fact a special case of VLAEs that admits efficient computation. In addition, we illustrate an example for Bernoulli output networks. 

In summary, the contributions of this work are as follows:
\begin{itemize}
\item We relate Gaussian output ReLU networks to probabilistic PCA \cite{tipping99} and thereby derive an iterative update for finding the mode of the posterior and a full-covariance Gaussian posterior approximation at the mode. 
\item We present \textit{Variational Laplace Autoencoders} (VLAEs), a general framework for training deep generative models based on the \textit{Laplace approximation} of the latent variable posterior. VLAEs not only minimize the amortization error but also provide the expressive power of full-covariance Gaussian, with no additional parameter overhead for the inference model. To the best of our knowledge, this work is the first attempt to apply the Laplace approximation to the training of deep generative models. 
\item We evaluate our approach on five benchmark datasets of MNIST, %\cite{lecun98}, 
Omniglot, %\cite{lake13}, 
Fashion-MNIST, %\cite{xiao17}, 
SVHN and %\cite{netzer11} and 
CIFAR-10. %\cite{krizhevsky09}. 
Empirical results show that VLAEs bring significant improvement over other recent amortized or iterative approaches, including VAE \cite{kingma14}, semi-amortized VAE \cite{kim18, marino18, krishnan18} and VAE with Householder Flow \cite{tomczak16}.
\end{itemize}

\section{Background}
\label{sec:background}
% can merge this into section 3

\subsection{Variational Autoencoders}
\label{sec:vae}
For the latent variable model $p_{\vtheta}(\x, \z) = p_{\vtheta}(\x|\z) p(\z)$, variational inference (VI) \cite{hinton1993, waterhouse1996, jordan1999} fits an approximate distribution $q(\z; \vct{\lambda})$ to the intractable posterior $p_{\vtheta}(\z|\x)$ and maximizes the \textit{evidence lower-bound} (ELBO):
\begin{align}
\mathcal{L}_{\vtheta}(\x; \vct{\lambda})
&= \mathbb{E}_{q(\z; \vct{\lambda})}[\ln p_{\vtheta}(\x, \z) - \ln q(\z; \vct{\lambda})] \\
&= \ln p_{\vtheta}(\x) - D_{KL}(q(\z; \vct{\lambda})||p_{\vtheta}(\z|\x)) \label{eq:elbo_gap} \\
&\le \ln p_{\vtheta}(\x),
\end{align}
where $q(\z; \vct{\lambda})$ is assumed to be a simple distribution such as diagonal Gaussian and $\vct{\lambda}$ is the variational parameter of the distribution (\textit{e.g.} the mean and covariance). 
The learning involves first finding the optimal variational parameter $\vct{\lambda}^*$ that minimizes the variational gap $D_{KL}(q(\z; \vct{\lambda})||p_{\vtheta}(\z|\x))$ between the ELBO and the true marginal log-likelihood, and then updating the generative model $\vct{\theta}$ using $\mathcal{L}_{\vtheta}(\x; \vct{\lambda}^*)$.

Variational autoencoders (VAE) \cite{kingma14} amortize the optimization problem of $\vct{\lambda}$ using the inference model $\vphi$ that dynamically predicts the approximate posterior as a function of $\x$:
\begin{align}
\mathcal{L}_{\vtheta, \vphi}(\x)
&= \mathbb{E}_{q_{\vphi}(\z|\x)}[\ln p_{\vtheta}(\x, \z) - \ln q_{\vphi}(\z|\x)]. \label{eq:elbo} %\\
%&= \ln p_{\vtheta}(\x) - D_{KL}(q_{\vphi}(\z|\x)||p_{\vtheta}(\z|\x)) \\
%&\le \ln p_{\vtheta}(\x).
\end{align}
The generative model (\textit{decoder}) and the inference model (\textit{encoder}) are jointly optimized. 
While such amortized variational inference (AVI) is highly efficient, there remain two fundamental challenges: (1) the limited expressiveness of the approximate posterior and (2) the amortization error, both of which will be discussed in the following sections. 

\subsection{Limited Posterior Expressiveness}
\label{sec:limited_express}
VAEs approximate the posterior with the fully-factorized Gaussian $q_{\vphi}(\z|\x) = \mathcal{N}(\vct{\mu}_{\vphi}(\x), \text{diag}(\vct{\sigma}^2_{\vphi}(\x))$. 
However the fully-factorized Gaussian may fail to accurately capture the complex true posterior distribution, causing the \textit{approximation error} \citep{cremer18}. 
As the ELBO (Eq.(\ref{eq:elbo})) tries to reduce the gap $D_{KL}(q_{\vphi}(\z|\x)||p_{\vtheta}(\z|\x))$,
it will force the true posterior $p_{\vtheta}(\z|\x)$ to match the fully-factorized Gaussian $q_{\vphi}(\z|\x)$, which negatively affect the capacity of the generative model \cite{mescheder17}. 

Hence, it is encouraged to use more expressive families of distributions; for example, one natural expansion is to model the full-covariance matrix $\mat{\Sigma}$ of the Gaussian. However, it requires $O(D^2)$ variational parameters to be predicted, placing a heavy burden on the inference model. %  

Normalizing flows \cite{rezende15, kingma16, tomczak16} approach this issue by using a class of invertible transformations whose densities are relatively easy to compute. 
However, the flow-based methods have drawbacks in that they incur additional parameter overhead for the inference model and are prone to the amortization error described below.

\subsection{The Amortization Error}
\label{sec:amortization_error}
Another problem of VAEs stems from the nature of amortized inference where the variational parameter $\vct{\lambda}$ is not explicitly optimized, but is dynamically predicted by the inference model. The error of the dynamic inference is referred to as the \textit{amortization error} and is closely related to the performance of the generative model \cite{cremer18}. 
% This amortization error can be quantified as $\mathcal{L}_{\vtheta}(\x; \vct{\lambda}^*) - \mathcal{L}_{\vtheta, \vphi}(\x)$ \citep{cremer18}. 
The suboptimal posterior predictions loosen the bound in the ELBO (Eq.(\ref{eq:elbo_gap})) and result in biased gradient signals flowing to the generative parameters. 
%\begin{align}
%\mathcal{L}_{\vtheta}(\x; \vct{\lambda}^*) - \mathcal{L}_{\vtheta, \vphi}(\x) 
%&= D_{KL}(q_{\vphi}(\z|\x)||p_{\vtheta}(\z|\x)) \\ 
%&- D_{KL}(q(\z; \vct{\lambda}^*)||p_{\vtheta}(\z|\x)). \nonumber
%\end{align}

Recently, \citet{kim18} and \citet{marino18} address this issue by iteratively updating the predicted variational parameters using gradient-based optimization.
However, they still rely on the fully-factorized Gaussian assumption, limiting the expressive power of the posterior.

\begin{figure}[t]
\vskip 0.05in
\begin{center}
\centerline{\includegraphics[width=0.8\columnwidth]{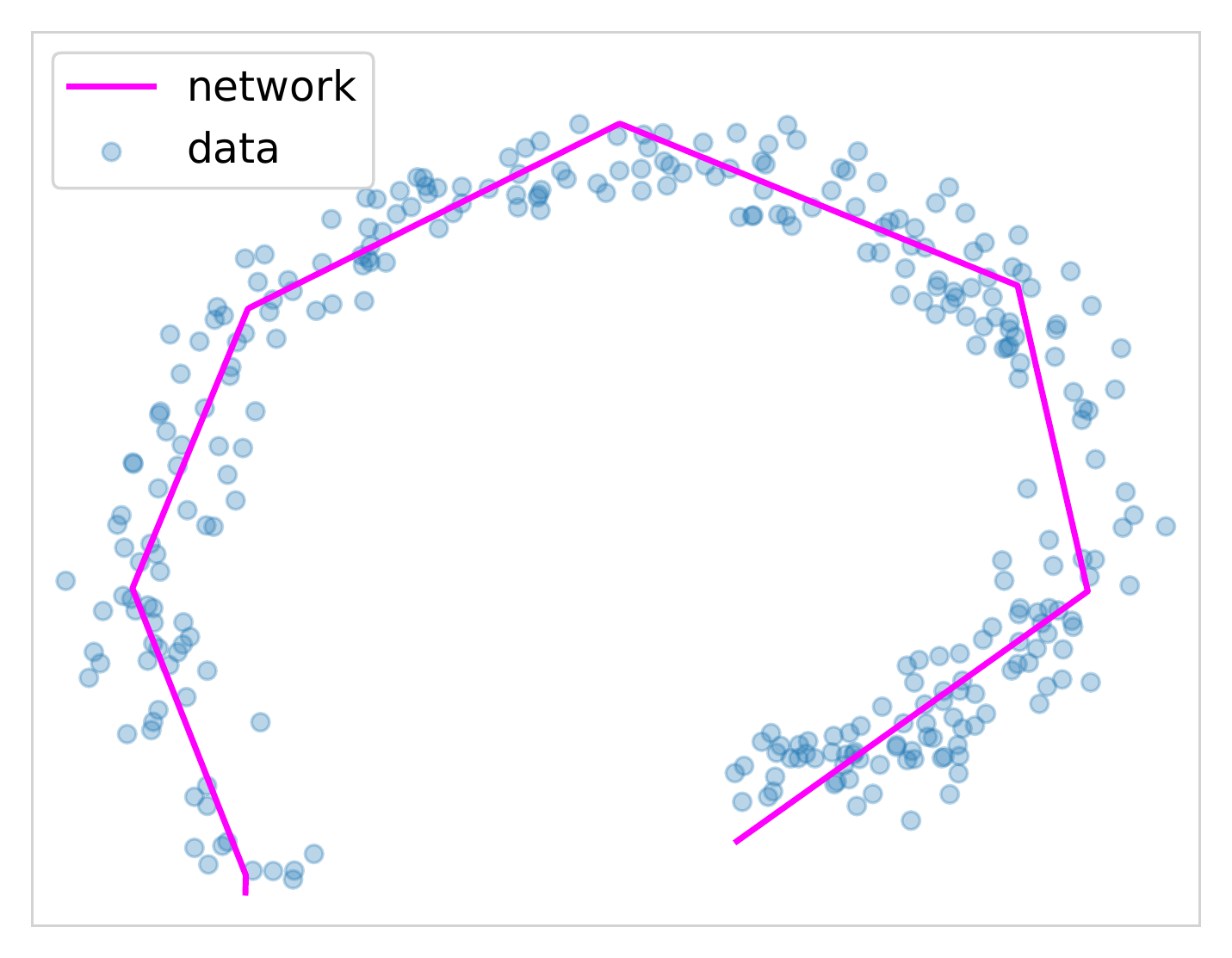}}
\caption{VAE's manifold learned on 2D toy data using a 1D latent variable. The VAE with rectified linear activation learns a piece-wise linear manifold and locally performs probabilistic PCA.}
\label{fig:toy}
\end{center}
\vskip -0.2in
\end{figure}

\section{Approach}
\label{sec:proposal}

We first describe probabilistic PCA (section \ref{sec:ppca}) and piece-wise linear ReLU networks (section \ref{sec:plnn}). Based on the local linearity of such networks, we derive an iterative approach to find the mode of the posterior and define a full-covariance Gaussian posterior at the mode (section \ref{sec:iterative_inference}). Finally, we present a general framework for training deep generative models using the Laplace approximation of the latent variable posterior (section \ref{sec:vlae}). 

\subsection{Probabilistic PCA}
\label{sec:ppca}
Probabilistic PCA \cite{tipping99} relates latent variable $\z$ to data $\x$ through linear mapping $\W$ as
\begin{align}
p(\z) &= \mathcal{N}(\vct{0}, \I), \\
p_{\vtheta}(\x|\z) &= \mathcal{N}(\mat{W} \z + \vct{b}, \sigma^2 \I),
\end{align}
where $\mat{W}, \vct{b}, \sigma$ are the parameters to be learned. 
Note that this is basically a \textit{linear} version of VAEs.
Under this particular model, the posterior distribution of $\z$ given $\x$ can be computed in a closed form \cite{tipping99}:
\begin{align}
p_{\vtheta}(\z|\x) &= \mathcal{N}(\frac{1}{\sigma^2} \mat{\Sigma} \W^T (\x - \vct{b}), \mat{\Sigma}), \label{eq:pca}\\
\text{ where } \mat{\Sigma} &= (\frac{1}{\sigma^2} \W^T \W + \I)^{-1}. \label{eq:pca_sigma}
\end{align}
%Note that the covariance matrix $\mat{\Sigma}$ depends on the linear mapping $\W$ and output variance $\sigma^2$. 

\subsection{Piece-wise Linear Neural Networks}
\label{sec:plnn}
Consider a following ReLU network $\vct{y} = g_{\vtheta}(\z)$:
\begin{alignat}{3}
&\h_{l+1} &&= \relu(\W_{l} \h_{l} + \vct{b}_{l}), \text{ for } l=0, \dots, L-1 \\
&\vct{y} &&= \W_L \h_L + \vct{b}_L, \label{eq:relu_network}
\end{alignat}
where $\h_{0} = \z$, $\relu(\x) = \max(\vct{0}, \x)$ and $L$ is the number of layers.

Our motivation is based on the observation that neural networks of rectified linear activations (\eg ReLU, Leaky ReLU, Maxout) are \textit{piece-wise linear} \cite{pascanu14, montufar14}. That is, the network segments the input space into linear regions within which it locally behaves as a linear function:
\begin{align}
g_{\vtheta}(\z+\vct{\epsilon}) \approx \W_{\z} (\z + \vct{\epsilon}) + \vct{b}_{\z}, \label{eq:local_linearity}
\end{align}
where the subscripts denote the dependence on $\z$.

To see this, note that applying a ReLU activation is equivalent to multiplying a corresponding mask matrix $\mat{O}$:
\begin{align}
\relu(\W \x + \vct{b}) = \mat{O} (\W \x + \vct{b}) \label{eq:relu_mask},
\end{align}
where $\mat{O}$ is a diagonal matrix whose diagonal element $o_i$ defines the activation pattern \cite{pascanu14}:
\begin{align}
o_{i} = 
\begin{cases}
1 \qquad \text{ if } \vct{w}_i^T \x + b_i > 0, \\
0 \qquad \text{ otherwise. }
\end{cases} \label{eq:relu_mask2}
\end{align}

The set of input $\x$ that satisfies the activation pattern $\{o_i\}_{i=1}^d$ such that $\{\x \,|\, \mathbb{I}(\vct{w}_i^T \x + b_i > 0) = o_i, \text{ for } i=1, \dots, d\}$ defines a convex polytope as it is an intersection of half-spaces. Accordingly, within this convex polytope the mask matrix $\mat{O}$ is constant.

We can obtain $\W_{\z}$ and $\vct{b}_{\z}$ in Eq.(\ref{eq:local_linearity}) by computing the activation masks during the forward pass and recursively multiplying them with the network weights:
\begin{align}
\vct{y}
&= \W_L \relu(\W_{L-1} \h_{L-1} + \vct{b}_{L-1}) + \vct{b}_L \label{eq:linear_map_start} \\
&= \W_L \mat{O}_{L-1} (\W_{L-1} \h_{L-1} + \vct{b}_{L-1}) + \vct{b}_L \\
&= \W_L \mat{O}_{L-1} \W_{L-1} \cdots \mat{O}_{0} \W_{0} \z + \dots \label{eq:linear_map1} \\
&= \mat{W}_{\z} \z + \vct{b}_{\z}. \label{eq:linear_map2}
\end{align}
Note that $\W_{\z}$ is the Jacobian of the network $\partial g_{\vtheta}(\z) / \partial \z$. Similar results apply to other kinds of piece-wise linear activations such as Leaky ReLU~\cite{maas13} and MaxOut~\cite{goodfellow13}.

\subsection{Posterior Inference for Piece-wise Linear Networks}
\label{sec:iterative_inference}
Consider the following nonlinear latent generative model:
\begin{align}
p(\z) &= \mathcal{N}(\vct{0}, \I), \label{eq:model1}\\
p_{\vtheta}(\x|\z) &= \mathcal{N}(g_{\vtheta}(\z), \sigma^2 \I), \label{eq:model2}
\end{align}
where $g_{\vtheta}(\z)$ is the ReLU network. In general, such nonlinear model does not allow the analytical computation of the posterior. Instead of using the amortized prediction $q_{\vphi}(\z|\x) = \mathcal{N}(\vct{\mu}_{\vphi}(\x), \text{diag}(\vct{\sigma}^2_{\vphi}(\x))$ like VAEs, we present a novel approach that exploits the piece-wise linearity of generative networks.  

The results in the previous section hints that the ReLU network $g_{\vtheta}(\z)$ learns the piece-wise linear manifold of the data as illustrated in Fig.\ref{fig:toy}, meaning that the model is locally equivalent to the probabilistic PCA. Based on this observation, we propose a new approach for posterior approximation which consists of two parts: (1) find the mode of posterior where probability density is mostly concentrated, and (2) apply local linear approximation of the generative network (Eq.(\ref{eq:local_linearity})) at the mode and analytically compute the posterior using the results of probabilistic PCA (Eq.(\ref{eq:pca})). Algorithm \ref{alg:inference} outlines the proposed method.

\begin{algorithm}[t]
	\caption{Posterior inference for piece-wise linear nets}
	\label{alg:inference}
	\begin{algorithmic}
		\STATE {\bfseries Input:} data $\x$, piece-wise linear generative network $g_{\vtheta}$, \\ 
		inference network $\text{enc}_{\vphi}$, update steps $T$, decay $\alpha_t$
		\STATE {\bfseries Output:} full-covariance Gaussian posterior $q(\z|\x)$
%		\STATE Sample $\x \sim p_{\text{data}}(\x)$
		\STATE $\vct{\mu}_0 = \text{enc}_{\vphi}(\x)$
		\FOR{$t=0$ {\bfseries to} $T-1$} 
		\STATE Compute linear approximation $g_{\vtheta}(\z) \approx \W_{t} \z + \vct{b}_{t}$
		% using Eq.(\ref{eq:linear_map_start}-\ref{eq:linear_map2})
		\STATE $\mat{\Sigma}_t \leftarrow (\sigma^{-2} \W_t^T \W_t + \I)^{-1}$
		\STATE $\vct{\mu}' \leftarrow \sigma^{-2} \mat{\Sigma}_{t} \W_{t}^T (\x - \vct{b}_{t})$
		\STATE $\vct{\mu}_{t+1} \leftarrow (1 - \alpha_t) \vct{\mu}_t + \alpha_t \vct{\mu}'$
		\ENDFOR
		\STATE Compute linear approximation $g_{\vtheta}(\z) \approx \W_{T} \, \z + \vct{b}_{T}$
		\STATE $\mat{\Sigma}_T \leftarrow (\sigma^{-2} \W_T^T \W_T + \I)^{-1}$
		\STATE $q(\z|\x) \leftarrow \mathcal{N}(\vct{\mu}_T, \mat{\Sigma}_T)$
		\STATE {\bfseries Return} $q(\z|\x)$
	\end{algorithmic}
\end{algorithm}

We derive an update equation for the posterior mode exploiting the local linearity of ReLU networks. The results of probabilistic PCA (Eq.(\ref{eq:pca})) leads to the solution for the posterior mode under the linear model $\vct{y} = \W \z + \vct{b}$. Based on this insight, we first assume linear approximation to the generative network $g_{\vtheta}(\z) \approx \W_{t} \z + \vct{b}_{t}$ at the current estimate $\vct{\mu}_t$ at step $t$ and update our mode estimate using the solution (Eq.(\ref{eq:pca})) under this linear model:
\begin{align}
\vct{\mu}_{t+1} &= \frac{1}{\sigma^2} \mat{\Sigma}_{t} \W_{t}^T (\x - \vct{b}_{t}), \label{eq:update_equation}
%\text{ where } \mat{\Sigma}_{t} &= (\frac{1}{\sigma^2} \W_{t}^T \W_{t} + \I)^{-1},
\end{align}
where $\mat{\Sigma}_{t}$ is defined as in Eq.(\ref{eq:pca_sigma}).

\begin{figure}[t]
\begin{center}
\centerline{\includegraphics[width=\columnwidth]{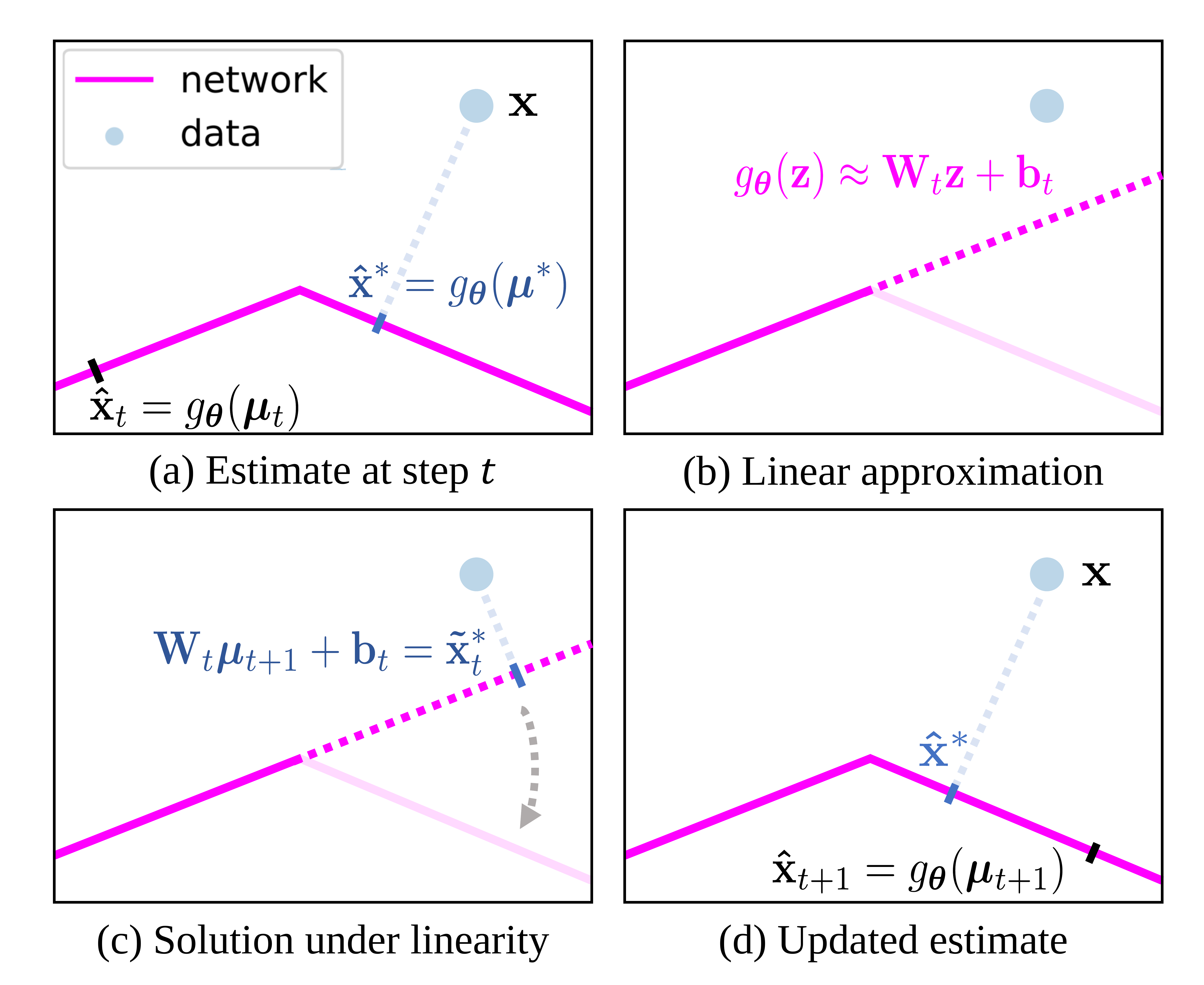}}
\caption{Illustration of the iterative update (Eq.(\ref{eq:update_equation}))\protect\footnotemark for the posterior mode, drawn on the data space. 
(a) The posterior mode $\vct{\mu}^*$ corresponds to the best reconstruction $\hat{\x}^*$ of data $\x$ on the network manifold, \ie $\hat{\x}^* = g_{\vtheta}(\vct{\mu}^*)$. 
$\hat{\x}_t$ shows the estimate at step $t$. 
%$\bar{\x}^*$ shows the optimal reconstruction. 
(b) We apply linear approximation to the network (dashed line).
%as in Eq.(\ref{eq:local_linearity}).
(c) We solve for $\vct{\mu}_{t+1}$ under this linear model (Eq.(\ref{eq:update_equation})). The network warps the result  according to its manifold (dotted arrow).
(d) Updated estimate. $\hat{\x}_{t+1}$ is now closer to $\hat{\x}^*$ than previous $\hat{\x}_t$. 
}
\label{fig:iterative_inference_toy}
\end{center}
\vskip -0.2in
\end{figure}

To take advantage of the efficiency of the amortized inference, we initialize the estimate using an inference model (encoder) as $\vct{\mu}_0 = \text{enc}_{\vphi}(\x)$ and iterate the update (Eq.(\ref{eq:update_equation})) for $T$ steps. Fig. \ref{fig:iterative_inference_toy} illustrates the process of iterative mode updates. 
We find that smoothing the update with decay $\alpha_t < 1$ improves the stability of the algorithm:	
\begin{align}
&\vct{\mu}_{t+1} = (1 - \alpha_t) \vct{\mu}_t + \alpha_t \vct{\mu}' \label{eq:update} \\
&\text{ where } \vct{\mu}' = \frac{1}{\sigma^2} \mat{\Sigma}_{t} \W_{t}^T (\x - \vct{b}_{t}).
%\mat{\Sigma}_{t} &= (\frac{1}{\sigma^2} \W_{t}^T \W_{t} + \I)^{-1}.
\end{align}

Finally, by assuming the linear model $g_{\vtheta}(\z) \approx \W_{T} \, \z + \vct{b}_{T}$ at $\vct{\mu}_T$, the approximate Gaussian posterior is defined:
\begin{align}
\label{eq:new_post_approx}
 q(\z|\x) &= \mathcal{N}(\vct{\mu}_T, \mat{\Sigma}_{T}), \\
\text{ where } \mat{\Sigma}_T &= (\frac{1}{\sigma^2} \W^T_T \W_T + \I)^{-1}.
\end{align}

We train our model by optimizing the ELBO in Eq.(\ref{eq:elbo}) by plugging in $q(\z|\x)$ of Eq.(\ref{eq:new_post_approx}). For sampling from the multivariate Gaussian and propagating the gradient to the inference model, we calculate the Cholesky decomposition $\mat{L} \mat{L}^T = \mat{\Sigma}_T$ and apply the reparameterization $\z = \vct{\mu} + \mat{L} \vct{\epsilon}$, where $\vct{\epsilon}$ is the standard Gaussian noise. 

\begin{figure}[t]
\begin{center}
\centerline{\includegraphics[width=0.8\columnwidth]{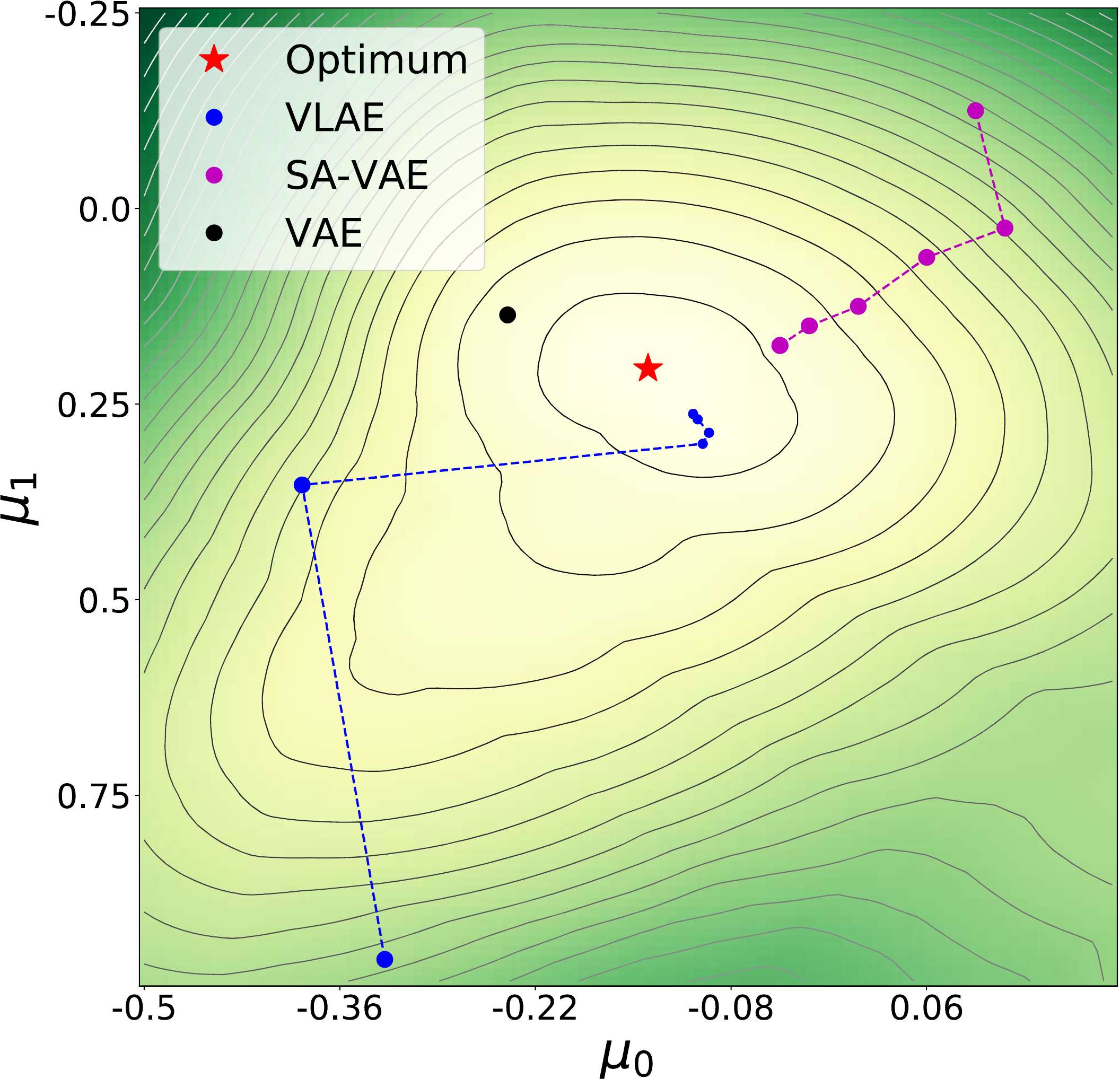}}
\caption{Illustration of the iterative mode update (Eq.(\ref{eq:update_equation})) on the ELBO landscape. The models are trained on MNIST using 2-dim latent variable. The update paths for the posterior mode $\vct{\mu}$ are depicted for five steps. The VLAE obtains the closest estimate, making large jumps during the process for faster convergence. 
%The optimum is found from grid search of the parameters. 
}
\label{fig:iterative_inference}
\end{center}
\vskip -0.2in
\end{figure}

\begin{figure}[h]
\begin{center}
\centerline{\includegraphics[width=\columnwidth]{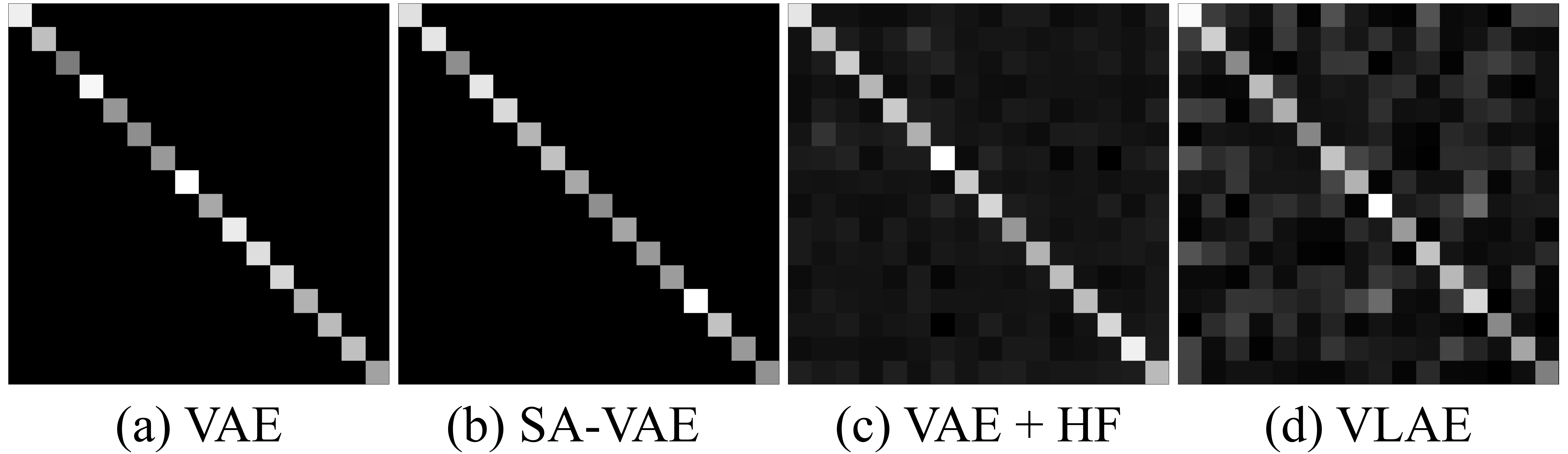}}
\caption{The covariance matrix of $q(\z|\x)$ from four different models. (a) VAE \citep{kingma14}. (b) Semi-Amortized VAE \cite{kim18}. (c) Householder Flow \cite{tomczak16}. (d) VLAE. While the former two only model diagonal elements, the latter two can model the full-covariance. The VLAE captures the richest correlations between dimensions.}
\label{fig:covariance}
\end{center}
\vskip -0.2in
\end{figure}

We highlight the notable characteristics of our approach:

\begin{itemize}
    \item We iteratively update the posterior mode rather than solely relying on the amortized prediction. This is in spirit similar to semi-amortized inference \cite{kim18,marino18,krishnan18}, but critical differences are: (1) our method can make large jumps than prevalent gradient-based methods by exploiting the local linearity of the network, and (2) it is efficient and deterministic since it require no sampling during updates. Fig. \ref{fig:iterative_inference} intuitively depicts these effects.
    \item We gain the expressiveness of the full-covariance Gaussian posterior (Fig \ref{fig:covariance})
%     whereas the semi-amortized inference methods 
	%\cite{kim18,marino18,krishnan18} 
%	are limited to the diagonal Gaussian assumption (Fig. \ref{fig:covariance}). 
	where the covariance is analytically computed from the local behavior of the generative network. 
	This is in contrast with normalizing flows \cite{rezende15, kingma16, tomczak16} which introduce extra parameter overhead for the inference model and hence are prone to the amortization error.
%	\item We do not update $\vct{\mu}_{t}$ and $\mat{\Sigma}_{t}$ to directly minimize $D_{KL}(q(\z|\x)||p_{\vtheta}(\z|\x))$, as opposed to standard variational approaches \cite{kim18,marino18,krishnan18}. However, empirical results in section \ref{sec:experiments} show that our approach significantly outperforms previous methods, thanks to the improved accuracy of iteratively updated posterior estimate and the expressive power of full-covariance Gaussian.  
\end{itemize}

\footnotetext{We here assume $\sigma^2 \rightarrow 0$ for the purpose of illustration. For $\sigma^2 > 0$, the prior shrinks the mode toward zero.}

\subsection{Variational Laplace Autoencoders}
\label{sec:vlae}
% In the following, we relate our approach to Laplace approximation and derive a generalization for general output distributions and activation functions. 
So far, the proposed approach assumes the neural networks with rectified linear activations and Gaussian output. We here present a general framework \textit{Variational Laplace Autoencoders} (VLAE), which are applicable to the general class of differentiable neural networks. To estimate the posterior $p_{\vtheta}(\z|\x)$, VLAEs employ the \textit{Laplace approximation} \cite{bishop06} that finds a Gaussian approximation based on the local curvature at the posterior mode.
We present how the Laplace method is incorporated for training deep generative models, and show the model in section \ref{sec:iterative_inference} is a special case of VLAEs where it admits efficient computations. 

Consider the problem of approximating the posterior $p_{\vtheta}(\z|\x)$ where we only have access to the unnormalized density $p_{\vtheta}(\x, \z)$. The Laplace method finds a Gaussian approximation $q(\z|\x)$ centered on the mode of $p_{\vtheta}(\z|\x)$ where the covariance is determined by the local curvature of $\log p_{\vtheta}(\x, \z)$. The overall procedure is largely divided into two parts: (1) finding the mode of posterior distribution and (2) computing the Gaussian approximation centered at the mode. 

First, we iteratively search for the mode $\vct{\mu}$ of $p_{\vtheta}(\z|\x)$ via
\begin{align}
\nabla_{\z} \log p_{\vtheta}(\x, \z) \rvert_{\z=\vct{\mu}} = \vct{0}. \label{eq:mode}
\end{align}
We can generally apply gradient-based optimization for this purpose.
After determining the mode, we run the second-order Taylor expansion centered at $\vct{\mu}$:
\begin{align}
&\log p_{\vtheta}(\x, \z) \approx \log p_{\vtheta}(\x, \vct{\mu}) - \frac{1}{2} (\z - \vct{\mu})^T \mat{\Lambda} (\z - \vct{\mu}), \\
&\text{ where } \mat{\Lambda} = - \nabla_{\z}^2 \log p_{\vtheta}(\x, \z) \rvert_{\z=\vct{\mu}}.
\end{align}
As this form is equivalent to Gaussian distribution, we define the approximate posterior as
\begin{align}
&q(\z|\x) = \mathcal{N}(\vct{\mu}, \mat{\Sigma}), \\
&\text{ where } \mat{\Sigma}^{-1} = \mat{\Lambda} = - \nabla_{\z}^2 \log p_{\vtheta}(\x, \z) \rvert_{\z=\vct{\mu}}.
\end{align}
This posterior distribution is then used to estimate the ELBO in Eq.(\ref{eq:elbo}) for training the generative model. Alg. \ref{alg:vlae} summarizes the proposed framework. 

\begin{algorithm}[tb]
	\caption{Variational Laplace Autoencoders}
	\label{alg:vlae}
	\begin{algorithmic}
		\STATE {\bfseries Input:} generative model $\vtheta$, inference model $\vphi$
		\STATE Sample $\x \sim p_{\text{data}}(\x)$
		\STATE Initialize $\vct{\mu}$ = $\text{enc}_{\vphi}(\x)$
		\FOR{$t=0$ {\bfseries to} $T-1$} 
		\STATE Update $\vct{\mu}$ (e.g. using gradient descent)
		\ENDFOR
		\STATE $\mat{\Sigma} \leftarrow (-\nabla^2_{\z} \ln_{\vtheta} p(\x, \z)|_{\z=\vct{\mu}})^{-1}$
		\STATE $q(\z|\x) \leftarrow \mathcal{N}(\vct{\mu}, \mat{\Sigma})$
		\STATE Sample $\vct{\epsilon} \sim \mathcal{N}(\vct{0}, \I)$
		\STATE Compute the Cholesky decomposition $\mat{L}\mat{L}^T = \mat{\Sigma}$
		\STATE $\z \leftarrow \vct{\mu} + \mat{L} \vct{\epsilon}$
		\STATE Estimate the ELBO: $\mathcal{L}_{\vtheta, \vphi}(\x)= \ln p_{\vtheta}(\x, \z) - \ln q_{\vphi}(\z|\x)$
		\STATE Update generative model: $\vtheta \leftarrow \vtheta + \alpha \nabla_{\vtheta} \mathcal{L}_{\vtheta, \vphi}(\x)$
		\STATE Update inference model: $\vphi \leftarrow \vphi + \alpha \nabla_{\vphi} \mathcal{L}_{\vtheta, \vphi}(\x)$
	\end{algorithmic}
\end{algorithm}

\textbf{Gaussian output ReLU networks}. We now make a connection to the approach in section \ref{sec:iterative_inference}. For the generative model defined in Eq.(\ref{eq:model1})--(\ref{eq:model2}), the joint log-likelihood is
\begin{align}
&\log p_{\vtheta}(\x, \z) \nonumber \\
&= -\frac{1}{2 \sigma^2}(\x - g_{\vtheta}(\z))^T (\x - g_{\vtheta}(\z)) - \frac{1}{2} \z^T \z + C,
\end{align}
where $C$ is a constant independent of $\z$. Taking the gradient with respect to $\z$ and setting it to zero,
\begin{align}
\nabla_{\z} \log p_{\vtheta}(\x, \z) = -\frac{1}{\sigma^2} \frac{\partial g_{\vtheta}(\z)^T}{\partial \z}  (g_{\vtheta}(\z) - \x) - \vct{z} = \vct{0}. \nonumber
\end{align}
By assuming linear approximation $g_{\vtheta}(\z) \approx \W_{\z} \z + \vct{b}_{\z}$ (Eq.(\ref{eq:local_linearity})) and plugging it in, the solution is
\begin{align}
\z = \frac{1}{\sigma^2} (\frac{1}{\sigma^2} \W_{\z}^T \W_{\z} + \I)^{-1} \W_{\z}^T (\x - \vct{b}_{\z}),
\end{align}
which is equivalent to the update equation of Eq.(\ref{eq:update_equation}).
Moreover, under the linear model above:
\begin{align}
\mat{\Lambda} = - \nabla_{\z}^2 \log p(\x, \z) &= (\frac{1}{\sigma^2} \W_{\z}^T \W_{\z} + \I), \label{eq:hessian_linear}
\end{align}
which agree with $\mat{\Sigma}^{-1}$ in Eq.(\ref{eq:pca}). That is, by using the local linearity assumption, we can estimate the covariance without the expensive computation of the Hessian of the generative network.

\textbf{Bernoulli output ReLU networks}. We illustrate an example for how VLAEs can be applied to Bernoulli output ReLU networks:
\begin{align}
&\log p_{\vtheta}(\x|\z) = \sum_i^n x_i \log y_i(\z) + (1 - x_i) \log (1 - y_i(\z)), \nonumber \\
&\text{ where }  \vct{y}(\z) = \frac{1}{1 + \exp(-g_{\vtheta}(\z))}.
\end{align}

Using $\nabla_{g_{\vtheta}(\z)} \log p(\x|\z) = \x - \vct{y}(\z)$ and chain rule, the gradient and the Hessian is
\begin{align}
&\nabla_{\z} \log p_{\vtheta}(\x, \z) = \frac{\partial g_{\vtheta}(\z)^T}{\partial \z} (\x -  \vct{y}(\z)) - \z, \\
&\nabla_{\z}^2 \log p_{\vtheta}(\x, \z) = \frac{\partial^2 g_{\vtheta}(\z)^T}{\partial \z^2} (\x - \vct{y}(\z)) \\
&- \frac{\partial g_{\vtheta}(\z)^T}{\partial \z} \text{diag}(\vct{y}(\z) \cdot (\vct{1} - \vct{y}(\z))) \frac{\partial g_{\vtheta}(\z)}{\partial \z} - \mat{I}. \nonumber
\end{align}
Plugging in the linear approximation $g_{\vtheta}(\z) \approx \W_{\z} \z + \vct{b}_{\z}$, the result simplifies to
\begin{align}
&\nabla_{\z} \log p_{\vtheta}(\x, \z) = \W_{\z}^T (\x -  \vct{y}(\z)) - \z, \label{eq:bernoulli_derivative}\\
&\nabla_{\z}^2 \log p_{\vtheta}(\x, \z) = -(\W_{\z}^T \mat{S} \W_{\z} + \mat{I}), \\
&\text{ where } \mat{S}_{\z} = \text{diag}(\vct{y}(\z) \cdot (\vct{1} - \vct{y}(\z))).
\end{align}
To solve Eq.(\ref{eq:bernoulli_derivative}), we first apply the first-order approximation of $\vct{y}(\z)$:
\begin{align}
\vct{y}(\z') 
&\approx \vct{y}(\z) + \frac{\partial \vct{y}(\z)}{\partial \z} (\z' - \z) \\
&= \vct{y}(\z) + \mat{S}_{\z} \W_{\z} (\z' - \z).
\end{align}
We plug it into Eq.(\ref{eq:bernoulli_derivative}) and solve the equation for zero, % leads to
\begin{align}
\z' = 
&(\W_{\z}^T \mat{S}_{\z} \W_{\z} + I)^{-1} \W_{\z}^T (\x - \vct{y}(\z) + \mat{S}_{\z} \W_{\z} \z). \nonumber
\end{align}
This leads to the update equation for the mode similar to the Gaussian case (Eq.(\ref{eq:update_equation})): 
\begin{align}
&\vct{\mu}_{t+1} = \mat{\Sigma}_{t} \W_{t}^T (\x - \vct{b}_{t}), \label{eq:update_equation_bernoulli} \\
&\text{ where } \mat{\Sigma}_{t} = (\W_{t}^T \mat{S}_{t} \W_{t} + \mat{I})^{-1}, 
\vct{b}_t = (\vct{y}_t - \mat{S}_{t} \W_{t} \vct{\mu}_t). \nonumber
\end{align}
After $T$ updates, the approximate posterior distribution for the Bernoulli output distribution is defined as
\begin{align}
q(\z|\x) &= \mathcal{N}(\vct{\mu}_T, \mat{\Sigma}_{T}).
\end{align}

\subsection{Efficient Computation}
\label{sec:computation}
For a network with width $O(D)$, the calculation of $\W_{\z}$ (Eq.(\ref{eq:linear_map1})--(\ref{eq:linear_map2})) and the update equation (Eq.(\ref{eq:update_equation})) requires a series of  matrix-matrix multiplication and matrix inversion of $O(D^3)$ complexity, whereas the standard forward or backward propagation through the network takes a series of matrix-vector multiplication of $O(D^2)$ cost. 
%This computation is amenable to GPU acceleration but is expensive compared to standard forward or backward propagation through the network, which takes a series of matrix-vector multiplication with $O(D^2)$ cost. 
As this computation 
%of update equation (Eq.(\ref{eq:update_equation})) or covariance matrix (Eq.(\ref{eq:pca_sigma})) 
can be burdensome for bigger networks\footnote{In our experiments, the computational overhead is affordable with GPU acceleration. For example, our MNIST model takes 10 GPU hours for 1,000 epochs on one Titan X Pascal.}, we discuss efficient alternatives for: (1) iterative mode seeking and (2) covariance estimation. 

\textbf{Iterative mode seeking}.
For piece-wise linear networks, nonlinear variants of Conjugate Gradient (CG) can be effective as the CG solves a system of linear equations efficiently. It requires the evaluation of matrix-vector products $\W_{\z} \z$ and $\W_{\z}^T \vct{r}$, where $\vct{r}$ is the residual $\x - (\W_{\z} \z + \vct{b})$. These are computable during the forward and backward pass with $O(D^2)$ complexity per iteration. See Appendix for details. 

For general differentiable neural networks, the gradient-based optimizers such as SGD, momentum or ADAM \cite{kingma15} can be adopted to find the solution of Eq.(\ref{eq:mode}). The backpropagation through the gradient-based updates requires the evaluation of Hessian-vector products but there are efficient approximations such as the finite differences \citep{lecun1993automatic} used in \citet{kim18}. 

\textbf{Covariance estimation}. 
Instead of analytically calculating the precision matrix $\mat{\Lambda} = \mat{\Sigma}^{-1} = \sigma^{-2} \W^T \W + \I$ (Eq.(\ref{eq:hessian_linear})) we may directly approximate the precision matrix using  truncated SVD for top $k$ singular values and vectors of $\mat{\Lambda}$. 
%It is equivalent to the best $k$-rank approximation of $\mat{\Lambda}$ in terms of Frobenius norm. 
The truncated SVD can be iteratively performed using the power method where each iteration involves evaluation of the Jacobian-vector product $\W^T \W \z$. This can be computed through the forward and backward propagation at $O(D^2)$ cost. One way to further accelerate the convergence of the power iterations is to extend the amortized inference model to predict $k$ vectors as seed vectors for the power method. However, we still need to compute the Cholesky decomposition of the covariance matrix for sampling.

\section{Related Work}
\label{related_works}
%Variational Autoencoders ~\cite{kingma} have a few known problems. Such as the posterior collapse issue, w
\citet{cremer18} and \citet{krishnan18} reveal that VAEs suffer from the inference gap between the ELBO and the marginal log-likelihood. 
\citet{cremer18} decompose this gap as the sum of the \textit{approximation error} and the \textit{amortization error}.  
The approximation error results from the choice of a particular variational family, such as fully-factorized Gaussians, restricting the distribution to be factorial or more technically, have a diagonal covariance matrix.
On the other hand, the amortization error is caused by the suboptimality of variational parameters due to the amortized predictions.

This work is most akin to the line of works that contribute to reducing the amortization error by iteratively updating the variational parameters to improve the approximate posterior \cite{kim18,marino18,krishnan18}. 
Distinct from the prior approaches which rely on gradient-based optimization, our method explores the local linearity of the network to make more efficient updates.

Regarding the approximation error, various approaches have been proposed for improved expressiveness of posterior approximation. Importance weighted autoencoders \cite{burda15} learn flexible posteriors using importance weighting. \citet{tran16} incorporate Gaussian processes to enrich posterior representation. \citet{maaloe16} augment the model with auxiliary variables and \citet{salimans15} use Markov chains with Hamiltonian dynamics.
Normalizing flows \cite{tabak13, rezende15, kingma16, tomczak16} transform a simple initial distribution to an increasingly flexible one by using a series of \textit{flows}, invertible transformations whose determinant of the Jacobian is easy to compute. For example, the Householder flows \cite{tomczak16} can represent a Gaussian distribution with a  full covariance alike to our VLAEs. However, as opposed to the flow-based approaches, VLAEs introduce no additional parameters and are robust to the amortization error. 

The Laplace approximation have been applied to estimate the uncertainty of weight parameters of neural networks \citep{mackay1992practical, ritter2018scalable}. However, due to the high dimensionality of neural network parameters (often over millions), strong assumptions on the structure of the Hessian matrix are required to make the computation feasible \citep{lecun1990optimal, ritter2018scalable}. On the other hand, the latent dimension of deep generative models is typically in  the hundreds, rendering our approach practical. 

\section{Experiments}
\label{sec:experiments}

We evaluate our approach on five popular datasets: MNIST \cite{lecun98}, Omniglot \cite{lake13}, Fashion-MNIST \cite{xiao17}, Street View House Numbers (SVHN) \cite{netzer11} and CIFAR-10 \cite{krizhevsky09}. 
We verify the effectiveness of our approach not only on the ReLU networks with Gaussian output (section \ref{sec:exp_gaussian}), but also on the Bernoulli output ReLU networks in section \ref{sec:vlae} on dynamically binarized MNIST (section \ref{sec:exp_bernoulli}). 

We compare the VLAE with other recent VAE models, including 
(1) VAE \cite{kingma14} using the standard fully-factorized Gaussian assumption, (2) Semi-Amortized VAE (SA-VAE) which extends the VAE using gradient-based updates of variational parameters \cite{kim18, marino18, krishnan18} (3) VAE augmented with Householder Flow (VAE+HF) \cite{tomczak16} \cite{rezende15} which employs a series of Householder transformation to model the covariance of the Gaussian posterior. For bigger networks, we also include (4) VAE augmented with Inverse Autoregressive Flow (VAE+IAF) \cite{kingma16}.

We experiment two network settings: (1) a small network with one hidden layer. The latent variable dimension is 16 and the hidden layer dimension is 256. We double both dimensions for color datasets of SVHN and CIFAR10. (2) A bigger network with two hidden layers. The latent variable dimension is 50 and the hidden layer dimension is 500 for all datasets. For both settings, we apply ReLU activation to hidden layers and use the same architecture for the encoder and decoder. See Appendix for more experimental details. 

The code is public at {\color{magenta}{http://vision.snu.ac.kr/projects/VLAE}}.

\subsection{Results of Gaussian Outputs}
\label{sec:exp_gaussian}

\begin{figure}[t]
\vskip 0.1in
\begin{center}
\centerline{\includegraphics[width=\columnwidth]{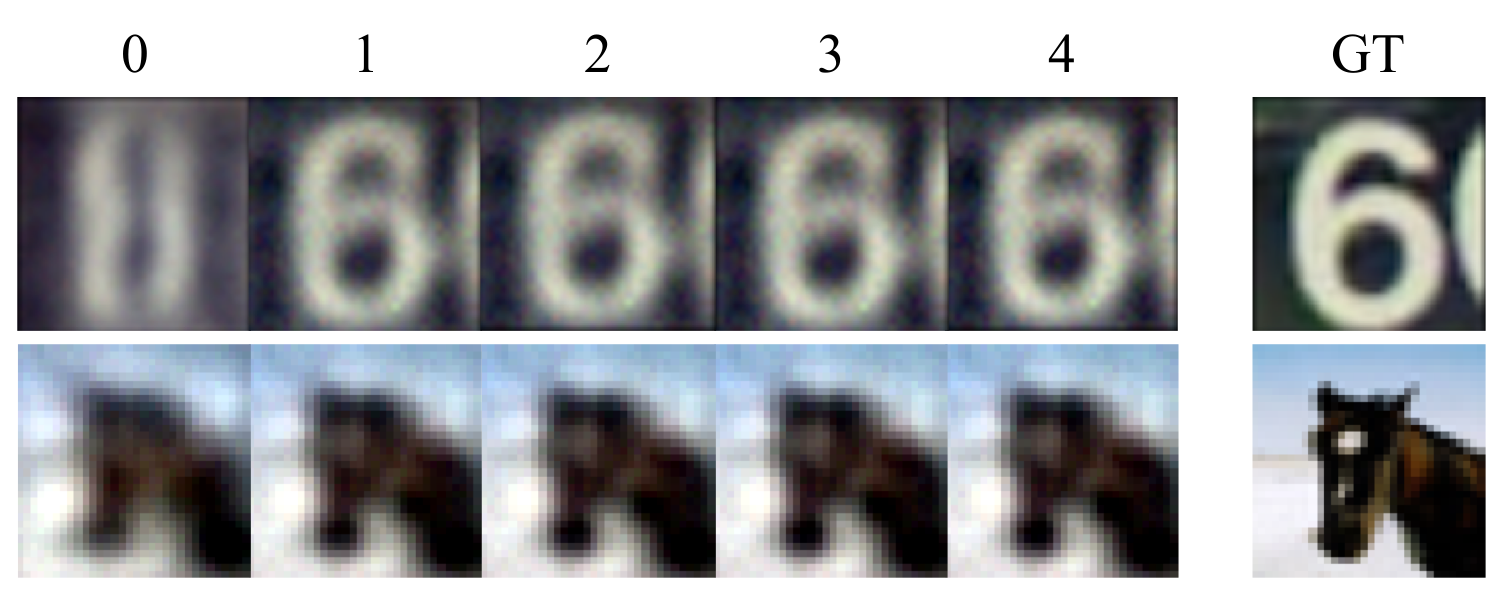}}
\caption{Examples of images reconstructed by VLAE with five update iterations proceeding from left to right. The pixel correlations greatly improve on the first update. The ground-truth samples are shown on the rightmost.}
\label{fig:recon_image}
\end{center}
\vskip -0.3in
\end{figure}

Table~\ref{tab:results} summarizes the results on the Gaussian output ReLU networks in the small network setting. The VLAE outperforms other baselines with notable margins in all the datasets, proving the effectiveness of our approach. Remarkably, a single step of update ($T=1$) leads to substantial improvement compared to the other models, and with more updates the performance further enhances. Fig. \ref{fig:recon_image} depicts how data reconstructions improve with the update steps. 

Table \ref{tab:results_big} shows the results with the bigger networks, which  bring considerable improvements. The VLAE again attains the best results for all the datasets. The VAE+IAF is generally strong among the baselines whereas the SA-VAE is worse compared to others. One distinguishing trend compared to the small network results is that increasing the number of updates $T$ often degrade the performance of the models. We suspect the increased depth due to the large number of updates or flows causes optimization difficulties, as we observe worse results on the training set as well.

For both network settings, the SA-VAE and VAE+HF show mixed results. We hypothesize the causes are as follows: 
(1) The SA-VAE update is noisy as the gradient of ELBO (Eq.(\ref{eq:elbo})) is estimated using a single sample of $\z$. Hence, it may cause instability during training and may require more iterations than used in our experiments for better performance. 
(2) Although the VAE+HF is endowed with flexibility to represent correlations between the latent variable dimensions, it is prone to suffer from the amortization error as it completely relies on the dynamic prediction of the inference network, whereas the inference problem becomes more complex with the enhanced flexibility. 

On the other hand, we argue that the VLAE is able to make significant improvements in fewer steps because the VLAE update is more powerful and deterministic (Eq.(\ref{eq:update_equation})), thanks to the local linearity of ReLU networks. Moreover, the VLAE computes the covariance of latent variables directly from the generative model, thus providing greater expressiveness with less amortization error. 

\begin{table}[t!]
	\centering
	\small
	\setlength{\tabcolsep}{2pt}
	\caption{Log-likelihood results on small networks, estimated with 100 importance samples. Gaussian output is used except the last column with the Bernoulli output. $T$ refers to the number of updates for the VLAE and SA-VAE \cite{kim18} or the number of flows for the VAE+HF \cite{tomczak16}.} % \cite{kim18} \cite{tomczak16}
	\label{tab:results}
	\vskip 0.15in
	
	\begin{sc}
		\begin{tabular}{lcccccc}
			\toprule
			% & \multicolumn{5}{c}{$Gaussian$} &  $Bernoulli$ \\
            % \cmidrule(lr){2-6}\cmidrule(lr){7-7}
			& \multirow{2}{*}{mnist}	& omni- & fashion & \multirow{2}{*}{svhn} & \multirow{2}{*}{cifar\scriptsize10} &  binary \\
			& 							& glot	& mnist	 & & & mnist   \\
			\midrule
			VAE    				& 612.9  & 343.5 	& 606.3 		&  	4555	& 2364 &   -96.73 \\
			SA-VAE	& & & & & &   \\
			\quad $T$=1		& 614.1  & 341.4	& 606.7 		& 	4553 	& 2366 &    -96.85 \\
			\quad $T$=2	  	& 615.2  & 346.6 	& 604.1			& 	4551	& 2366  &    -96.73\\
			\quad $T$=4   	& 612.8  & 348.6 	& 606.6 		&  	4553	& 2366  &    -96.71 \\
			\quad $T$=8   	& 612.1  & 345.5 	& 608.0 		&	4559	& 2365  &    -96.89\\
			VAE+HF	 & & & & & &   \\
			\quad $T$=1	& 610.5  & 341.5 	& 604.3 		&  	4557	& 2366  &    -96.75\\
			\quad $T$=2	& 613.1  & 343.1 	& 606.5 		&  	4569	& 2361 &     -96.52\\
			\quad $T$=4	& 612.9  & 333.8	& 604.9 		&  	4564	& 2362 &    -96.44\\
			\quad $T$=8	& 615.6  & 332.6 	& 605.5 		&  	4536	& 2357  &   -96.14\\
			\midrule
			VLAE & & & & & &   \\
			\quad $T$=1 		& 638.6  & 362.0 	& 614.9 		&  	4639	& 2374  &   -94.68\\
			\quad $T$=2 		& 645.4  & 372.7 	& 615.5 		&  	4681	& 2381  &   -94.46\\
			\quad $T$=4 		& 649.9  & 372.3 	& 615.6 		&  	4711	& 2387  &  \textbf{-94.41}\\
			\quad $T$=8 		& \textbf{650.3}  & \textbf{380.7} 	& \textbf{618.8} & \textbf{4718} & \textbf{2392}  &   -94.57 \\
			\bottomrule 
		\end{tabular}
	\end{sc}
	\vskip -0.2in
\end{table}

\subsection{Results of Bernoulli Outputs}
\label{sec:exp_bernoulli}

To show the generality of our approach on non-Gaussian output, we experiment ReLU networks with Bernoulli output on dynamically binarized MNIST. Each pixel is stochastically set to 1 or 0 with a probability proportional to the pixel intensity \cite{salakhutdinov08}. 

The rightmost columns in Table \ref{tab:results}--\ref{tab:results_big} show the results on dynamically binarized MNIST. The VLAE attains the highest log-likelihood as in the Gaussian output experiments. The results demonstrate that VLAEs are also effective for non-Gaussian output models.

\begin{table}[t!]
	\centering
	\small
	\setlength{\tabcolsep}{2pt}
	\caption{Log-likelihood results on bigger networks, estimated with 5000 importance samples. See the caption of Table \ref{tab:results} for details.}
	\label{tab:results_big}
	\vskip 0.15in
	\begin{sc}
		\begin{tabular}{lcccccc}
			\toprule
			& \multirow{2}{*}{mnist}	& omni- & fashion & \multirow{2}{*}{svhn} & \multirow{2}{*}{cifar\scriptsize10} &  binary \\
			& 							& glot	& mnist	 & & & mnist   \\

			\midrule
			VAE    				& 1015 	& 602.7 	& 707.7			& 5162		& 2640 		&  -85.38 \\
			SA-VAE & & & & & & \\
			\quad $T$=1		& 984.7 & 598.5		& 706.4 		& 5181 		& 2639 		&  -85.20 \\
			\quad $T$=2   	& 1006  & 589.8		& 708.3			& 5165		& 2639  	&  -85.10 \\
			\quad $T$=4   	& 999.8 & 604.6 	& 706.4			& 5172		& 2640  	&  -85.43 \\
			\quad $T$=8   	& 990.0 & 602.7 	& 697.0			& 5172		& 2639  	&  -85.24 \\
			VAE+HF & & & & & & \\
			\quad $T$=1		& 1028  & 602.5 	& 715.6 		& 5201		& 2637  	&  -85.27 \\
			\quad $T$=2		& 1020  & 603.1 	& 710.4 		& 5179		& 2636 		&  -85.31 \\
			\quad $T$=4		& 989.8 & 607.0		& 710.0 		& 5209		& 2641 		&  -85.22 \\
			\quad $T$=8		& 944.3 & 608.5		& 714.6 		& 5196		& 2640  	&  -85.41 \\
			VAE+IAF & & & & & & \\
			\quad $T$=1		& 1015	& 609.4		& 721.2			& 5037 		& 2642 		&  -84.26 \\
			\quad $T$=2		& 1057	& 617.5		& 724.1			& 5150 		& 2624 		&  -84.16 \\
			\quad $T$=4		& 1051  & 617.5		& 721.0			& 4994 		& 2638 		&  -84.03 \\
			\quad $T$=8		& 1018	& 606.7		& 725.7			& 4951 		& 2639 		&  -83.80 \\
			\midrule
			VLAE & & & & & & \\
            \quad $T$=1 		& \textbf{1150}	& 727.4 	& 817.7 		&  5324  & \textbf{2687}  &  -83.72 \\
			\quad $T$=2 		& 1096	& \textbf{731.1}    & 825.4            &  5159  & 2686 		&  -83.84 \\
			\quad $T$=4 		& 1054 	& 701.2 	        & \textbf{826.0}        &  5231  & 2683  	&  -83.73 \\
			\quad $T$=8 		& 1009 	& 661.2 	        & 821.0                 & \textbf{5341} &  2639  &  \textbf{-83.60} \\
			\bottomrule
		\end{tabular}
	\end{sc}
	\vskip -0.2in
\end{table}

\section{Conclusion}
\label{sec:conclusion}
We presented \textit{Variational Laplace Autoencoders} (VLAEs), which apply the Laplace approximation of the posterior for training deep generative models. The iterative mode updates and full-covariance Gaussian approximation using the curvature of the generative network enhances the expressive power of the posterior with less amortization error.
The experiments demonstrated that on ReLU networks, the VLAEs outperformed other amortized or iterative models. 

As future work, an important study may be to extend VLAEs to deep latent models based on the combination of top-down information and bottom-up inference \cite{kingma16, sonderby16}.  

\section*{Acknowledgements}

This work is supported by Samsung Advanced Institute of Technology, Korea-U.K. FP Programme through NRF of Korea (NRF-2017K1A3A1A16067245)
and IITP grant funded by the Korea government (MSIP) (2019-0-01082).

\bibliography{icml19_vi}
\bibliographystyle{icml2019}

%%%%%%%%%%%%%%%%%%%%%%%%%%%%%%%%%%%%%%%%%%%%%%%%%%%%%%%%%%%%%%%%%%%%%%%%%%%%%%%
%%%%%%%%%%%%%%%%%%%%%%%%%%%%%%%%%%%%%%%%%%%%%%%%%%%%%%%%%%%%%%%%%%%%%%%%%%%%%%%
% DELETE THIS PART. DO NOT PLACE CONTENT AFTER THE REFERENCES!
%%%%%%%%%%%%%%%%%%%%%%%%%%%%%%%%%%%%%%%%%%%%%%%%%%%%%%%%%%%%%%%%%%%%%%%%%%%%%%%
%%%%%%%%%%%%%%%%%%%%%%%%%%%%%%%%%%%%%%%%%%%%%%%%%%%%%%%%%%%%%%%%%%%%%%%%%%%%%%%

\end{document}

\twocolumn[
\icmltitle{[Supplementary] \\ Variational Laplace Autoencoders}
\icmlsetsymbol{equal}{*}

\begin{icmlauthorlist}
\icmlauthor{Yookoon Park}{to}
\icmlauthor{Chris Dongjoo Kim}{to}
\icmlauthor{Gunhee Kim}{to}
\end{icmlauthorlist}

\icmlaffiliation{to}{Department of Computer Science, Seoul National University, Seoul, South Korea}

\icmlcorrespondingauthor{Gunhee Kim}{gunhee@snu.ac.kr}

% You may provide any keywords that you
% find helpful for describing your paper; these are used to populate
% the "keywords" metadata in the PDF but will not be shown in the document
\icmlkeywords{Variational Autoencoders, Variational Inferece, Deep Generative Models, Amortization Gap, Approximation Gap}

\vskip 0.3in
]

\begin{figure*}[t]
	\centering
	\begin{subfigure}{0.3\textwidth}
		\includegraphics[width=\textwidth]{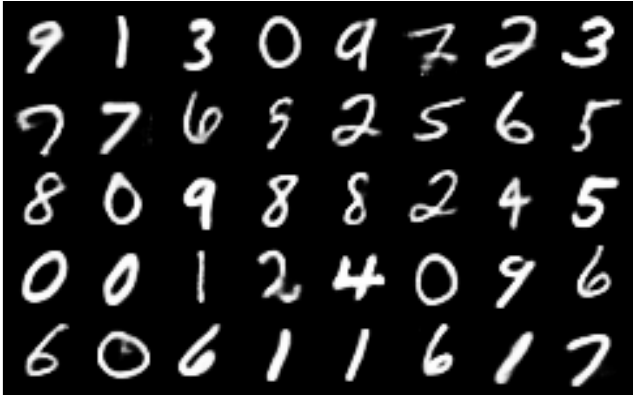}
		\caption{MNIST (Bernoulli)}
	\end{subfigure}
	\begin{subfigure}{0.3\textwidth}
		\includegraphics[width=\textwidth]{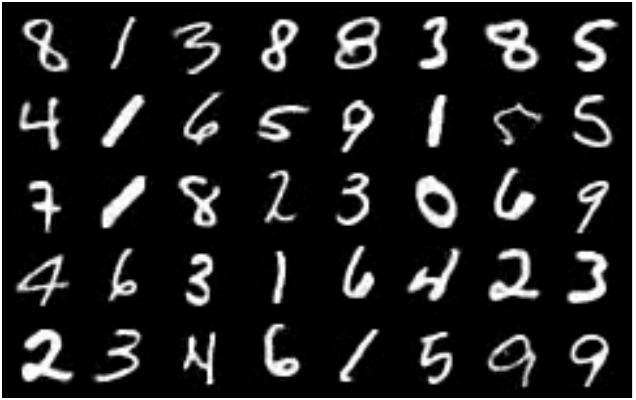}
		\caption{MNIST}
	\end{subfigure}
	\begin{subfigure}{0.3\textwidth}
		\includegraphics[width=\textwidth]{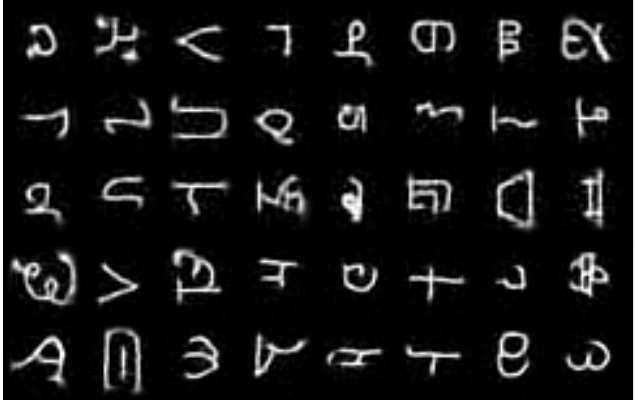}
		\caption{Omniglot}
	\end{subfigure}
	\\\begin{subfigure}{0.3\textwidth}
		\includegraphics[width=\textwidth]{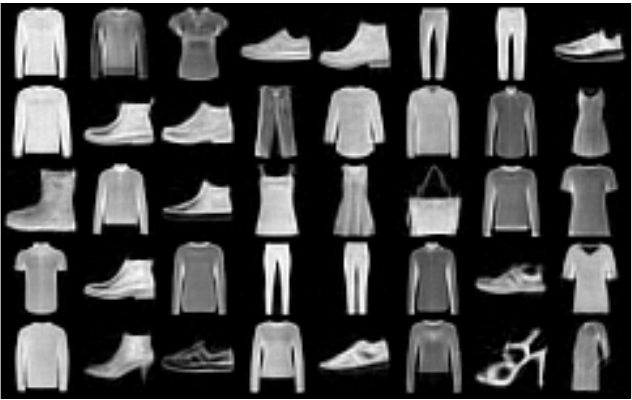}
		\caption{Fashion MNIST}
	\end{subfigure}
	\begin{subfigure}{0.3\textwidth}
		\includegraphics[width=\textwidth]{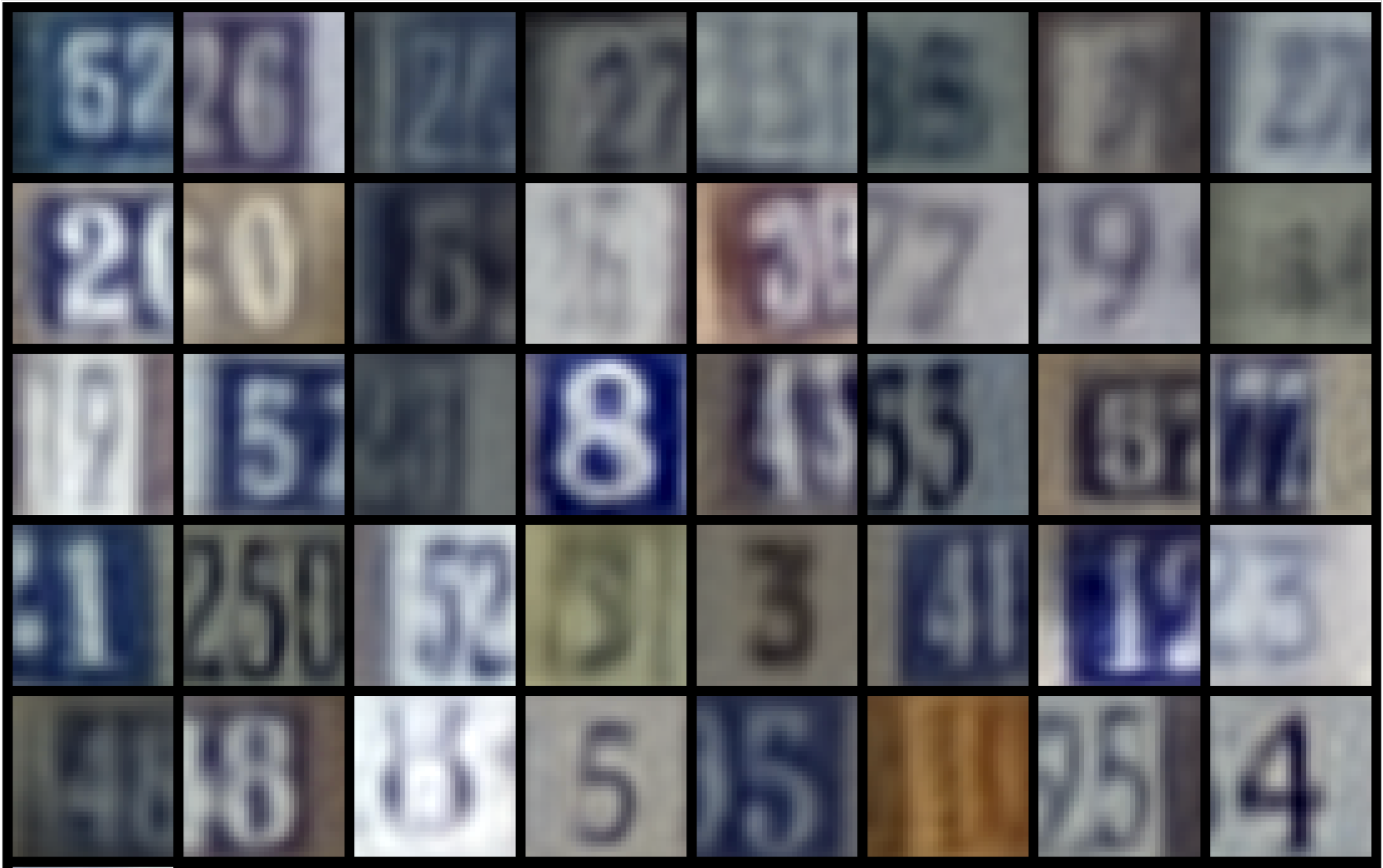}
		\caption{SVHN}
	\end{subfigure}
	\begin{subfigure}{0.3\textwidth}
		\includegraphics[width=\textwidth]{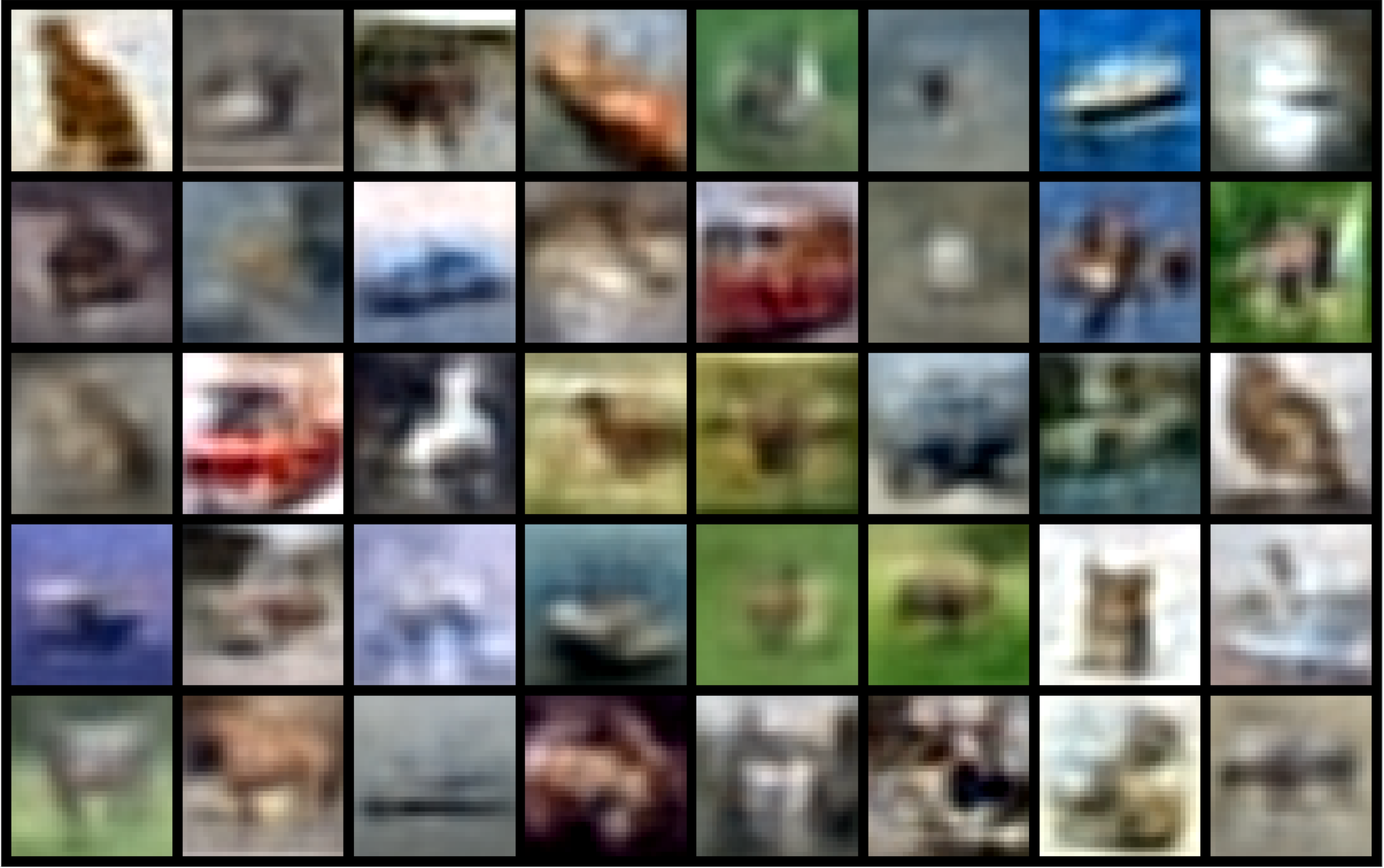}
		\caption{CIFAR10}
	\end{subfigure}
	\caption{Examples of sample reconstructions by VLAE. Gaussian output distribution is used unless otherwise stated.}
	\label{fig:reconstruction}
\end{figure*}

\begin{figure*}[h!]
	\centering
	\begin{subfigure}{0.3\textwidth}
		\includegraphics[width=\textwidth]{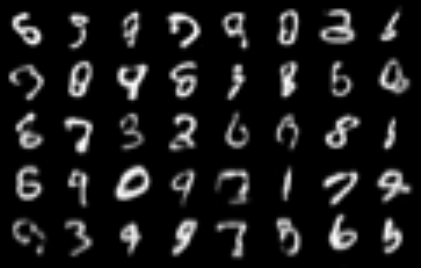}
		\caption{MNIST (Bernoulli)}
	\end{subfigure}
	\begin{subfigure}{0.3\textwidth}
		\includegraphics[width=\textwidth]{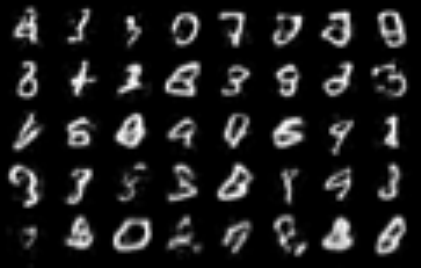}
		\caption{MNIST}
	\end{subfigure}
	\begin{subfigure}{0.3\textwidth}
		\includegraphics[width=\textwidth]{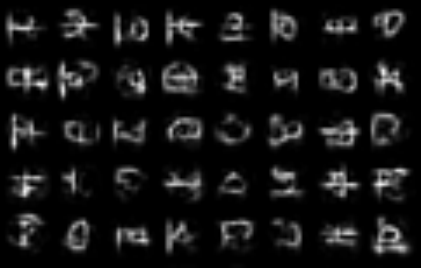}
		\caption{Omniglot}
	\end{subfigure}
	\\\begin{subfigure}{0.3\textwidth}
		\includegraphics[width=\textwidth]{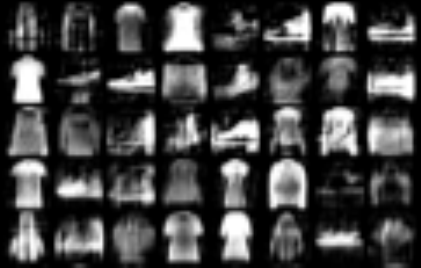}
		\caption{Fashion MNIST}
	\end{subfigure}
	\begin{subfigure}{0.3\textwidth}
		\includegraphics[width=\textwidth]{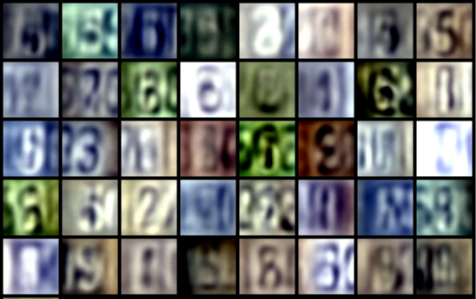}
		\caption{SVHN}
	\end{subfigure}
	\begin{subfigure}{0.3\textwidth}
		\includegraphics[width=\textwidth]{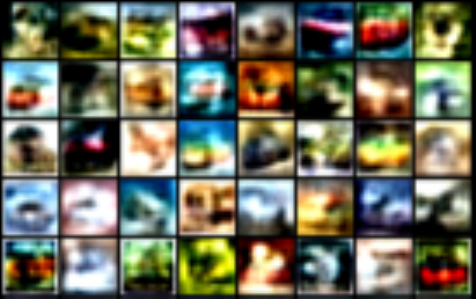}
		\caption{CIFAR10}
	\end{subfigure}
	\caption{Examples of output samples generated by VLAE. Gaussian output distribution is used unless otherwise stated.}
	\label{fig:generated}
\end{figure*}

\section{Image Samples}
\label{sec:generation_outputs}

Figure \ref{fig:reconstruction} and \ref{fig:generated} illustrates examples of reconstruction and generation samples made by the VLAE on MNIST \cite{lecun98}, Fashion MNIST \cite{xiao17}, Omniglot \cite{lake13}, SVHN \cite{netzer11} and CIFAR 10 \cite{krizhevsky09}. Overall, the reconstructions are sharp but generated samples tend to be blurry, especially when the data is complex (e.g. CIFAR10). We expect using convolutional architectures to be helpful for improving image generation qualities.

\section{Experimental Details}
\label{sec:exp_details}
We optimize using ADAM \cite{kingma15} with learning rate 0.0005. Other parameters of the optimizer is set to default values. We experiment with $T=1, 2, 4, 8$ where $T$ is the number of iterative updates for VLAE and SA-VAE, or the number of flow transformations for VAE+HF. 
We set the batch size to 128. All models are trained up to 2000 epochs at maximum and evaluated using the checkpoint that gives the best validation performance. 

We set $\alpha_t = 0.5 / (t+1)$ as decay for the VLAE update. For SA-VAE, the variational parameter $\vct{\lambda}_t$ is updated $T$ times using SGD: $\vct{\lambda}_{t+1} = \vct{\lambda}_t + \alpha \frac{\partial}{\partial \vphi} \mathcal{L}_{\vtheta}(\x; \vct{\lambda}_t)$ with $\alpha=0.0005$. The value of $\alpha$ is determined using a grid search among \{1.0, 0.1, 0.001, 0.0005, 0.0001\} on the small network. We estimate the gradient using a single sample of $\z$ and apply the gradient norm clipping to avoid divergence of SA-VAE.

In our experiments, we find that the bigger models are susceptible to the parameter initialization, and their latent variables are prone to collapse if not properly initialized.
We also observe VLAE and VAE+IAF are relatively robust to hyperparameter settings compared to other models.
In order to prevent latent variable collapse, we use He's initialization \citep{he15} to preserve variance of backward propagation with gain of $2^{1/3}$ to account for the network structure that consists of two ReLU layer and one linear layer. In this way, the variance of gradients is preserved in initial phase of training. Furthermore, the data is mean-normalized and scaled so that the reconstruction loss at initial state is approximately $1$. 
These changes successfully prevent latent variable collapse and significantly improve overall performance of the models. 

This finding hints that it is crucial to preserve gradient variance throughout the networks. 
Note that the gradient signal to the encoder comes from two sources: (1) KL divergence term $D_{KL}(q_{\vphi}(\z|\x)||p(\z))$ (2) Reconstruction term $\mathbb{E}_{q_{\vphi}(\z|\x)}[\ln p_{\vtheta}(\x|\z)]$. While the gradient from the KL divergence term is directly fed into the encoder, the gradient from the reconstruction term - which is essential for preventing the latent variable collapse - have to propagate backwards through the decoder to reach the encoder. We hypothesize that if the networks are not initialized properly, the reconstruction gradient is overwhelmed by the KL divergence gradient which drives the approximate posterior $q_{\vphi}(\z|\x)$ to collapse to the prior $p(\z)$ in the initial stage of training. 

\section{Conjugate Gradient Method}
To measure the performance of the Conjugate Gradient (CG) method as an alternative to the update equations of the main draft, we implement the VLAE+CG model where the mode update equation is replaced with the CG ascent step. 
We use nonlinear Conjugate Gradient of Polak-Ribi\'ere method. For more details on nonlinear Conjugate Gradient methods, we refer readers to \citep{shewchuk1994introduction, dai2010nonlinear}. With $T=4$, we find that VLAE+CG yields about 25\% speed-up compared to the VLAE with minor performance loss ($\sim$ 1\%) on MNIST. 

\bibliography{icml19_vi_supp}
\bibliographystyle{icml2019}